# Challenges in Developing LRs for Non-Scheduled Languages: A Case of Magahi

**Ritesh Kumar**     **Bornini Lahiri**     **Deepak Alok**

Centre for Linguistics
Jawaharlal Nehru University, New Delhi, India
{riteshkrjnu, lahiri.bornini, deepakalok06}@gmail.com

**Abstract**
Magahi is an Indo-Aryan Language, spoken mainly in the Eastern parts of India. Despite having a significant number of speakers, there has been virtually no language resource (LR) or language technology (LT) developed for the language, mainly because of its status as a non-scheduled language. The present paper describes an attempt to develop an annotated corpus of Magahi. The data is mainly taken from a couple of blogs in Magahi, some collection of stories in Magahi and the recordings of conversation in Magahi and it is annotated at the POS level using BIS tagset.

## 1. Introduction

Grierson (1903) has classified Magahi under Eastern group of Outer sub-branch of Indo-Aryan languages. Scholars like Turner have clubbed the 'Bihari' languages with Eastern and Western Hindi (Masica, 1991). Chatterji (1926) has given an entirely different classification where Western Hindi is almost an isolated group while Eastern Hindi, Bihari and other languages of Eastern group are clubbed together. But the classification given above by Grierson (1903) is most widely accepted one and a similar classification is given by Jeffers (1976) and others.

Magahi is spoken mainly in Eastern states of India including Bihar and Jharkhand, along with some parts of West Bengal and Orissa. The variety of Magahi spoken in and around Gaya, Jehanabad and Patna is generally considered standard.

### 1.1. Linguistic Features of Magahi

Magahi, like other Eastern Indo-Aryan and unlike other Indo-Aryan languages, including Hindi, do not have number and gender agreement. It has only person and honorificity agreement with the verb (Verma, 1991). It is a nominative-accusative, highly inflected language with almost free order of constituents within phrases and sentences. Both adjectives and nouns in Magahi have two basic forms, which may be loosely termed as form 1 and form 2. For example, $g^hora$ 'horse', *sona* 'gold', *ujər* 'white', etc are the form 1 while $g^hor$-*ba*, *son-ma, ujər-ka* are their respective form 2. The suffixes attached to these words may be considered some kind of affixal particles with different kinds of linguistic functions (like specificity/definiteness/focus, etc.)(Alok,2010). Magahi also has a couple of numeral classifiers, *ɡo* and *tʰo*. Two of the most unique features of Magahi is simultaneous agreement of verb with subject as well as direct object and also agreement of verb with the addressee, who is not present at the time of the utterance (Verma, 1991 and Verma, 2003).

### 1.2. Socio-political Situation of Magahi

Although linguistically Magahi is not even part of the same sub-family as Hindi (as is clear from the Grierson's classification) and they have very different grammars, politically as well as socially Magahi is labeled as a dialect of Hindi (so always clubbed with Hindi in any kind of Government schemes or funding for the languages). Constitutionally also being a non-scheduled language (Constitution of India has a Eighth Schedule which has 22 major languages listed in it and are considered scheduled/official languages of the country), Magahi (along with several other languages of India) is a completely ignored language. It is not the official medium of instruction in any school and neither the language itself nor its literature studied anywhere in the formal education system. This has further resulted in the depletion of the use of language for literary and other such purposes which involve writing. At present, Magahi has more than a few volumes of written literature. Along with this a couple of quarterly journals are published in Magahi. Besides this some books on Magahi literature and grammar and linguistic studies on Magahi written in Hindi or English are available.

However despite this attitude of the Indian government and a comparatively scant availability of written materials, a large population (counting up to 13.978,565 according to Census of India, 2001) residing in the villages and small towns still speak and use the language as their medium of communication and information-sharing. Consequently it becomes very necessary that the language resources (LRs) and language technologies (LTs) be developed for Magahi also (separately from the dominant language, Hindi), which is expected to have an impact at two different levels – one, it may affect the overall attitude towards the language and inculcate more positive attitude towards it, two, it will open a whole new world of information and knowledge (which could be disseminated with the help of the technology) to millions of people in a language in which they are most comfortable and familiar with.

As of now very little has been done towards the development of language technology for Magahi. The only notable exception, to the best of my knowledge is the 'Magahi Verbs Project' done by Ritesh Kumar, under the supervision of Dr. Girish Nath Jha (hosted at http://sanskrit.jnu.ac.in/student_projects/magahi-search.jsp ). It is a very rudimentary tool which gives the analysis of a Magahi verb form and also generates the rest of the forms of that verb.

In this paper we present the first steps in an attempt to develop a deeply annotated corpus of Magahi, which could prove to be very useful for the development of other resources like lexical databases, dictionaries, etc and also more sophisticated language technologies for Magahi. It is also expected to serve as a model and inspiration for the development of LR and LT for other non-scheduled and smaller languages of India.

## 2. Compiling the Corpus

### 2.1. Data Sources

Considering the fact that Magahi does not have a vast collection of written literature, getting data for the corpus was a big challenge. The major part of the corpus, till now, is taken from the following sources -

a. The two blogs (give the names). These are the only blogs which I could find in Magahi. So in order to have a representation of the kind of language used on the Internet, the data from these two blogs were included. Out of these two, one consists of the texts taken from different sources like collection of stories and texts on Magahi grammar, while the other one contains the original jottings of the writer.

b. Some volumes of two Magahi magazines – Magadhbhumi and Alkaa Maagadhi

c. A Collection of Magahi folktales

d. Transcript of the recordings of the spontaneous conversation in Magahi. The recordings have been carried out both in private, home domain as well as in the public domain. The recordings mainly includes from the speakers of the Central Magahi variety.

As of now, altogether a corpus of around 1 million words has been prepared.

### 2.2. Structure of the Corpus

The corpus is divided into three subcorpora -

a. Magahi in indirect written communication – It basically includes the data taken from the literary texts and other written texts. They are not meant to communicate anything *directly* to the readers in the same way as conversation; while in conversation a reciprocal is expected and considered appropriate, in these written texts, it is generally a one-way communication with very little answer expected from the readers. This subcorpora is further divided into two subcorpora -

1. Data from the Magahi magazines
2. Data from Magahi books

Both of these are further divided into three different classes -

1. Data from the prosaic literary texts (includes short stories and novels)
2. Data from dramatic literary texts (includes full length plays and one-act plays)
2. Data from the poetic literary texts
3. Data from the non-fictional texts

b. Magahi in direct written communication (chiefly, Computer-mediated Communication here) – It includes the data taken from the computer-mediated communication as well as in some other forms of written conversation. It is further divided into two subcorpora -

1. Data from Magahi used in Computer-mediated communication (CMC), which is again divided into two classes -
   i. Synchronous CMC
   ii. Asynchronous CMC
2. Non-CMC Magahi Data – It includes data chiefly from the letters and is divided into two domains -
   i. Personal Domain – Letters shared among family members, relatives, etc.
   ii. Public Domain – Letters written to the newspapers, magazines, etc.

c. Magahi in spoken communication – It includes the transcripts of the recordings made in two domains -
   i. Personal Domain – Conversation among family members in the home.
   ii. Public Domain – Conversation among neighbours and fellow-villagers outside the home.

At present the data is arranged in several excel files; each file containing data from one source (for example, one story or one blog entry, etc.). Each sentence in the corpus is given a unique ID assigned to it. However it will be later exported to the standard XML database for an easier and more efficient use in NLP.

### 2.3. Preparing the Metadata

Preparation of metadata is one of the most significant steps in the creation of any corpus since it, one one hand, helps in arranging the data in a proper format, and on the other hand, it helps other people in making an optimum, informed and clear use of the corpus.

In the present corpus two separate metadata files are maintained for each subcorpora. One file contains what is called the *cataloguing* and *administrative* metadata by Austin (2006). The other file contains the other kinds of metadata described by Austin, including the *descriptive*, *structural* and *technical* metadata. The information contained in the cataloguing metadata file is as described below.

a. For the indirect written communication, the information like title of the chapter/story from which the data is taken, the name of its author, the name of the book/magazine, the date of publication, the page numbers from which the sentences are taken, the date on which the entry has been made in the corpus and the name of the person who has made the entry in corpus are included in the metadata. This information, along with the ID of the sentences related to this information, is maintained for each file in the corpus.

b. For the asynchronous CMC in direct written communication, the information like the name of the blog/web portal, the name of the writer, the date on which the entry was put up on the web, the date on which it was included in the corpus and the web link from which the entry was retrieved are included in the metadata. For synchronous CMC and non-CMC, only the date and the details about the participants are included in the metadata file.

c. For the transcripts of spoken communication, a detailed information regarding the recordings from which the data is taken (which includes the information regarding date, time and place of recordings, the person who has recorded the data, the original format and encoding details of recording, the format and encoding details of the recordings in the current state, the machine on which recordings were made, a small description of the context in which recording was made, the total duration of the recorded file from which the data is taken, the name of the software used, if any, for exporting data from the recorder to the computer, the people involved in the conversation and others present on the spot) is maintained.

The other metadata file contains a short summary of the kind of data found in the subcorpora, the theme of each file (descriptive metadata), a specification of the way data is arranged in the corpora (structural metadata) and also a description of the softwares that could be used to access the files (technical).

Besides these a general metadata file will be created for the whole corpora which could also be used to access the corpora. This will contain the information about location of different kinds of files, the subcorpora to which they belong, the statistical information (like the number of words and the sentences) and the hyperlinks leading to those files/folders.

## 3. Annotating the Corpus

It is very necessary to annotate the data to make it useful for NLP research. Considering the fact that it consists of both written and spoken data and it will also be used for the development of several applications for Magahi, it needs to be annotated at several different levels including POS, syntactic, semantic, discourse level, etc. in order to make its complete and efficient use. The POS annotation of the corpus has started and we also plan to annotate it at deeper levels.

POS annotation of the corpus is being carried out using the tagset derived out of the standard POS tagset for Indian Languages, designed, developed and recognised by the Bureau of Indian Standards (BIS). This tagset has been released in the year 2010 and then subsequently modified in 2011 and is now supposed to be followed as a national standard for the POS annotation of any data or corpus in any of the Indian languages. It is a hierarchical tagset prepared with inputs from two of the most popular tagsets for Indian languages - Microsoft Research India's Indian Languages Part of Speech Tagset (ILPOST) and the tagset of Indian Languages Machine Translation (ILMT) project. The general annotation scheme consists of two levels of categories. Top-level consists of 11 categories and includes nouns, verb, demonstrative, pronoun, etc. This category has a sub-category at the second level like proper noun, common noun, etc for the top-level category 'noun'. A tag has been assigned to categories at both the levels. While the tag of the lower level has to be assigned manually, the tag of the upper level is assigned automatically.

### 3.1. Preparing the Magahi Tagset

The Magahi tagset is derived and prepared out of this super-tagset approved by BIS tagset. The tagset is listed in Table 1 below. This tagset has a couple of differences from the Hindi tagset (even though Magahi is considered a dialect of Hindi). While Hindi does not have classifiers, Magahi has a couple of them; so it is added in the Magahi tagset. Moreover the verbal morphology of Magahi is significantly different from Magahi and so the information regarding finiteness (which could not be decided in Hindi at the word-level) is available at the word-level, thus, the two sub-categories of main verb – finite and non-finite are added in this tagset.

| Sl. No | Category | | | Label | Annotation Convention | Examples (in IPA) |
|---|---|---|---|---|---|---|
| | Top level | Subtype (level 1) | Subtype (level 2) | | | |
| 1 | **Noun** | | | N | N | cʰɔːɽɑ (boy) |
| 1.1 | | Common | | NN | N__NN | cəcəriː (a small bridge-like st.) ləŋgte (naked) |
| 1.2 | | Proper | | NNP | N__NNP | pʰuləva |
| 1.4 | | Nloc | | NST | N__NST | əgaɽiː, picʰaɽiː |
| 2 | **Pronoun** | | | **PR** | **PR** | |
| 2.1 | | Personal | | PRP | PR__PRP | həm, həməniː |
| 2.2 | | Reflexive | | PRF | PR__PRF | əpəne |
| 2.3 | | Relative | | PRL | PR__PRL | ɟe, ɟekər |
| | | Reciprocal | | PRC | PR__PRC | əpəne |
| 2.4 | | Wh-word | | PRQ | PR__PRQ | kɑ, ke |
| 2.5 | | Indefinite | | PRI | PR__PRI | koi, kekrɑ |
| **3** | **Demonstrative** | | | **DM** | **DM** | |
| 3.1 | | Deictic | | DMD | DM__DMD | ĩhã, ʊ̃hã |
| 3.2 | | Relative | | DMR | DM__DMR | ɟe, ɟəun |
| 3.3 | | Wh-word | | DMQ | DM__DMQ | kekrɑ, kəun |
| 3.4 | | Indefinite | | DMI | DM__DMI | i, ʊ |
| **4** | **Verb** | | | **V** | **V** | ləukna (to see) |
| 4.1 | | Main | | VM | V__VM | pʰĩcnɑ (to wash clothes) əɟʰurɑnɑ (to get entangled) |
| 4.1.1 | | | Finite | VF | V_VM__VF | goṭli: (soaked clothes), goṭbo |
| 4.1.2 | | | Non-finite | VNF | V_VM__VNF | goṭənɑ (to soak clothes) |
| 4.1.3 | | | Participle Noun | VNP | V_VM__VNP | goṭevɑlɑ (one who soak) |
| 4.2 | | Auxiliary | | VAUX | V__VAUX | həi, həli:, həṭʰi: |
| **5** | **Adjective** | | | **JJ** | **JJ** | cəkəitʰ (short and well-built) bəṭpʰəros (uselessly talkative) |
| **6** | **Adverb** | | | **RB** | **RB** | cəbʰak (with splash) cəbʰər-cəbʰr (a manner of eating) |
| **7** | **Postposition** | | | **PSP** | **PSP** | ke, me, pər, ɟore |
| **8** | **Conjunction** | | | **CC** | **CC** | |
| 8.1 | | Co-ordinator | | CCD | CC__CCD | ɑʊ, bɑki:, bəluk |
| 8.2 | | Subordinator | | CCS | CC__CCS | kɑheki, ṭə, ki |
| **9** | **Particles** | | | **RP** | **RP** | |
| 9.1 | | Default | | RPD | RP__RPD | ṭə, bʰi: |
| 9.2 | | Classifier | | CL | RP__CL | ɡo, tʰo |
| 9.3 | | Interjection | | INJ | RP__INJ | əre, he, cʰi:, bɑpre |
| 9.4 | | Intensifier | | INTF | RP__INTF | təhtəh, tuhtuh, bʰək-bʰək |
| 9.5 | | Negation | | NEG | RP__NEG | nə, məṭ, binɑ |
| **10** | **Quantifiers** | | | **QT** | **QT** | ek, pəhila, kucʰ |
| 10.1 | | General | | QTF | QT__QTF | təni:sun, dʰermən i: |
| 10.2 | | Cardinals | | QTC | QT__QTC | ek, du, iɡɑrəh |
| 10.3 | | Ordinals | | QTO | QT__QTO | pəhila, dʊsrɑ |
| **11** | **Residuals** | | | **RD** | **RD** | |
| 11.1 | | Foreign word | | RDF | RD__RDF | A word in foreign script. |

| 11.2 | | Symbol | | SYM | RD__SYM | For symbols such as $, & etc |
| 11.3 | | Punctuation | | PUNC | RD__PUNC | Only for punctuations |
| 11.4 | | Unknown | | UNK | RD__UNK | |
| 11.5 | | Echowords | | ECH | RD__ECH | (pɑni:-) uni: (kʰɑnɑ-) unɑ |

Table 1: The Magahi POS Tagset

This tagset is used to tag the corpora with POS information. Till now around 10,000 words in the corpus have been annotated with POS information.

### 3.2. The Annotation Tool: ILCIANN

ILCIANN (Indian Languages Corpora Initiative Annotation Tool) is an online annotation tool developed as part of Indian Language Corpora Initiative (ILCI Project) (Jha, 2010) for annotating the ILCI corpora. This tool is being used to annotate this corpus also. Some of the advantages of using such a tool includes –

a. No need of setting up any extra software or resources for tagging.
b. The tool autotags a lot of words (based on an autotag list), thereby, saving lots of effort and time
c. The annotated files are saved in an uniform and clean format (with no noise whatsoever) since it is saved by the tool itself in the required format, thereby, saving efforts and time on post-processing and data clean-up.

A snapshot of the tool is given in Fig. 1 below.

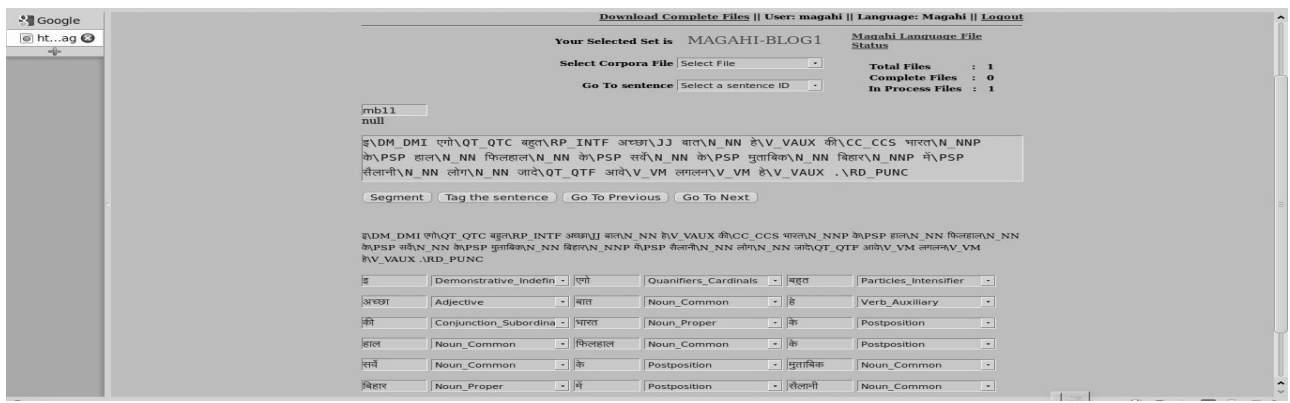

Fig. 1: A Snapshot of the ILCIANN Tool used for annotating the corpus Data

### 4. The Way Ahead

POS annotation is the most basic level of annotation in order to make the corpus useful for NLP. However in order to make the corpus more powerful and capable of being used for more complex tasks, it is very necessary to annotate it with lot more information. Moreover, along with this, the development of some basic technologies like POS tagger, morphological analysers and parsers are required so that higher-level applications could be developed for the language. After the POS tagging, the data will be further annotated with detailed morpho-syntactic and semantic information (using 'construction-labelling' method, suggested by Hellan, et al. (2009) for LRLs) as well as for the discourse-level information (by adapting DIT++ dialog act annotation scheme[1] for the present purposes).

---

[1] http://dit.uvt.nl/